\documentclass[11pt,twocolumn, clsnofoot]{IEEEtran}

\ifCLASSOPTIONcompsoc
\else
\fi
\ifCLASSINFOpdf
\else
\fi
\usepackage{graphicx}
\usepackage{subfigure}
\usepackage{multirow}%
\usepackage{cite}

\newtheorem{theorem}{Theorem}
\newtheorem{corollary}{Corollary}
\newtheorem{lemma}{Lemma}
\newtheorem{remark}{Remark}

\newtheorem{assumption}{Assumption}

\usepackage{algorithm}
\usepackage{algorithmic}

\usepackage{graphicx}
\usepackage{subfigure}

\begin{document}

\title{Re-scale boosting for regression and classification}

\author{Shaobo Lin, ~Yao Wang~, and Lin Xu % <-this % stops a space
\IEEEcompsocitemizethanks{\IEEEcompsocthanksitem S. Lin is with the
College of the Mathematics and Information Science, WenZhou
University, Wenzhou 325035, P R China. Y. Wang is with the
Department of Statistics, Xi'an Jiaotong University, Xi'an 710049, P
R China, and L. Xu is with the Institute for Information and System
Sciences, Xi'an Jiaotong University, Xi'an 710049, P R China.
Authors contributed equally to this paper and are listed
alphabetically.}}

\IEEEcompsoctitleabstractindextext{%
\begin{abstract}
%\boldmath
Boosting    is a learning scheme that combines weak prediction rules
to produce a strong composite estimator, with the underlying
intuition that one can obtain accurate prediction rules by combining
``rough'' ones. Although boosting is proved to be  consistent and
overfitting-resistant, its numerical convergence rate
  is relatively slow. The aim of this paper is to develop a new
boosting strategy, called the re-scale boosting (RBoosting), to
accelerate the numerical convergence rate and, consequently, improve
the learning performance of boosting. Our studies show that
RBoosting possesses the almost optimal numerical convergence rate in
the sense that, up to a logarithmic factor, it can reach the minimax
nonlinear approximation rate. We then use RBoosting to tackle both
the  classification and regression  problems,   and deduce a tight
generalization error estimate. The theoretical and experimental
results show that RBoosting outperforms boosting in terms of
generalization.

\end{abstract}

}

% make the title area
\maketitle

\IEEEdisplaynotcompsoctitleabstractindextext

\IEEEpeerreviewmaketitle

\section{Introduction}
Contemporary scientific investigations frequently encounter a common
issue of exploring the relationship between a response and a number
of covariates. Statistically, this issue can be usually modeled to
minimize either an empirical loss function or a penalized empirical
loss. Boosting  is recognized as a state-of-the-art scheme to attack
this issue and has triggered enormous research activities in the
past twenty years
\cite{Duffy2002,Freund1995,Friendman2001,Schapire1990}.

Boosting   is an iterative procedure that combines weak prediction
rules to produce a strong composite learner, with the underlying
intuition that one can obtain accurate prediction rules by combining
``rough'' ones. The gradient descent view \cite{Friendman2001} of
boosting shows that it can be regarded as a step-wise fitting scheme
of additive models. This statistical viewpoint connects various
boosting algorithms to   optimization problems with corresponding
loss functions. For example, $L_2$ boosting \cite{Buhlmann2003} can
be interpreted  as a stepwise learning scheme to the $L_2$ risk
minimization problem. Also, AdaBoost \cite{Freund1996} corresponds
to an approximate optimization of the exponential risk.

Although the success of the initial boosting algorithm (Algorithm
\ref{alg1} below) on many data sets and its ``resistance to
overfitting'' were comprehensively demonstrated
\cite{Buhlmann2003,Freund1996}, the problem is
 that its numerical convergence rate is usually a bit
slow \cite{Livshits2009}. In fact,   Livshits \cite{Livshits2009}
proved that for some sparse target functions, the numerical
convergence rate of boosting lies in
$(C_0k^{-0.1898},C_0'k^{-0.182})$, which is much slower than the
minimax nonlinear approximation rate $\mathcal O(k^{-1/2})$. Here
and hereafter, $k$ denotes the number of iterations, and $C_0,C_0'$
are absolute constants. Various modified versions of boosting have
been proposed to accelerate  its numerical convergence rate and then
to improve its generalization capability. Typical examples include
the regularized boosting via shrinkage (RSboosting)
\cite{Ehrlinger2012} that multiplies a small regularization factor
to the step-size deduced from the linear search, regularized
boosting via truncation (RTboosting) \cite{Zhang2005} which
truncates the linear search in a small interval and
$\varepsilon$-boosting \cite{Hastie2007} that specifies the
step-size as a fixed small positive number $\varepsilon$ rather than
using the linear search.

The purpose of the present paper is to propose a new  modification
of boosting to accelerate the numerical convergence rate  of
boosting to the near optimal  rate $\mathcal O(k^{-1/2}\log k)$ .
The new variant of boosting, called the re-scale boosting
(RBoosting), cheers the philosophy behind the faith ``no pain, no
gain'', that is, to derive the new estimator, we always take a
shrinkage operator to re-scale the old one. This idea is similar as
the ``greedy algorithm with free relaxation '' \cite{Temlyakov2012}
or ``sequential greedy algorithm'' \cite{Zhang2003} in sparse
approximation and is essentially different from Zhao and Yu's Blasso
\cite{Zhao2007}, since the shrinkage operator is imposed to the
composite estimator rather than the new selected weak learner. With
the help of the shrinkage operator, we can derive different types of
RBoosting such as the re-scale AdaBoost, re-scale Logitboost, and
re-scale $L_2$ boosting for regression and classification.

We present both theoretical analysis and experimental verification
to classify the performance of  RBoosting with convex loss
functions. The main contributions can be concluded as four aspects.
At first, we deduce the (near) optimal  numerical convergence rate
of RBoosting. Our result shows that RBoosting can  improve the
numerical convergence rate of boosting to the (near) optimal rate.
Secondly, we derive the generalization error bound of RBoosting. It
is shown that the generalization capability of RBoosting is
essentially better than that of boosting. Thirdly, we deduce the
consistency of RBoosting. The consistency of boosting has already
justified  in \cite{Bartlett2007} for AdaBoost. The novelty of our
result  is that the consistency of RBoosting is built upon relaxing
the restrictions to the dictionary and providing more flexible
choice of the iteration number. Finally, we experimentally compare
RBoosting with boosting, RTboosting, RSboosting and
$\varepsilon$-boosting  in both regression and classification
problems. Simulation results demonstrate that, similar to other
modified versions of boosting, RBoosting outperforms  boosting in
terms of prediction accuracy.

The rest of paper can be organized as follows. In Section 2, we
introduce RBoosting and compare it with other related algorithms. In
Section 3, we study the theoretical behaviors  of RBoosting, where
its numerical convergence, consistency and generalization error
bound are derived. In Section 4, we employ   a series of simulations
to verify our assertions. In the last section, we draw a simple
conclusion and present some further discussions.

\section{Re-scale boosting}

In classification or regression problems with a covariate or
predictor variable $X$ on $\mathcal X\subseteq\mathbf R^d$ and a
real response variable $Y$, we observe $m$ i.i.d. samples ${\bf
Z}^m=\{(X_1,Y_1),\dots,(X_m,Y_m)\}$ from an unknown  distribution
$D$. Consider a loss function $\phi(f,y)$ and define $Q(f)$ (true
risk) and   $ {Q}_m(f)$ (empirical risk) as
$$
         Q(f)=\mathbf E_D\phi(f(X),Y),
$$
and
$$
         {Q}_m(f)= {\mathbf E}_{\bf Z}\phi(f(X),Y)=\frac1m\sum_{i=1}^m\phi(f(X_i),Y_i),
$$
where $\mathbf E_D$ is the expectation over the unknown true joint
distribution $D$ of $(X,Y)$  and $ {\mathbf E}_{\bf Z}$ is the
empirical expectation based on the sample ${\bf Z}^m$.

Let $
          S=\{g_1,\dots,g_n\}
$
 be the set of weak learners (classifiers or regressors)  and define
$$
          \mbox{Span}(S)=\left\{\sum_{j=1}^na_jg_j:g_j\in S,
          a_j\in\mathbf R, n\in\mathbf N\right\}.
$$
We assume that $\phi$, therefore $Q_m$, is Fr\'{e}chet
differentiable and denote by $Q'_m(f,h)=(\nabla Q_m(f),h)$   the
value of linear functional $\nabla Q_m(f)$ at $h$, where $\nabla
Q_m(f)$ satisfies, for all $f,g\in\mbox{Span}(S)$,
$$
             \lim_{t\rightarrow0}\frac1t(Q_m(f+th)-Q_m(f))=(\nabla
             Q_m(f),h).
$$

Then the gradient descent view of boosting \cite{Friendman2001}  can
be interpreted as the following Algorithm \ref{alg1}.

\begin{algorithm}[H]\caption{Boosting}\label{alg1}
\begin{algorithmic}
\STATE {{ Step 1 (Initialization)}: Given data
$\{(X_i,Y_i):i=1,\dots,m\}$, weak learner set (or dictionary) $S$,
iteration number $k^*$, and $f_0\in\mbox{Span}(S)$}.
 \STATE{ { Step 2 (Projection of gradient
%\footnote{ To facilitate the use, it is preferable to
%introduce a nondecreasing auxiliary function $\psi$ \cite{Zhang2005}
%to construct a new functionals for  certain loss functions,
%$$
%          A_m(f)=\psi(Q_m(f))=\psi\left(\frac1m\sum_{i=1}^m\phi(f(X_i),Y_i)\right).
%$$
%}
)}: Find $g_k^*\in S$ such that
$$
              - Q_m'(f_{k-1},g_k^*)=\sup_{g\in S}-Q_m'(f_{k-1},g).
$$}
 \STATE{{Step 3 (Linear search)}:
 Find $\beta_k^*\in \mathbf R$  such that
$$
       Q_m(f_k+\beta_k^*g^*_k)= \inf_{\beta_k\in\mathbf
       R}Q_m(f_k+\beta_kg^*_k).
$$
Update $f_{k+1}=f_k+\beta_k^*g_k^*$.}
 \STATE{ { Step 4 (Iteration)}: Increase $k$ by one and repeat Step 2 and Step 3 if
$k<k^*$.}
\end{algorithmic}
\end{algorithm}

Although this original  boosting algorithm   was proved to be
consistent \cite{Bartlett2007} and overfitting resistant
\cite{Friendman2000}, a series of studies
\cite{DeVore1996,Livshits2009,Temlyakov2008a}   showed that its
numerical convergence rate is far slower than that of the best
nonlinear approximant. The main reason is that the linear search
  in Algorithm \ref{alg1} makes $f_{k}$ be not always the
greediest one. In particular, as shown in Fig.1, if $f_{k-1}$ walks
along the direction of $g_k$ to $\theta_0 g_k$, then  there usually
exists a weak learner $g$ such that the angle $\alpha=\beta$. That
is, after $\theta_0 g_k$, continuing to walk along $g_k$ is no more
the greediest one. However, the linear search makes $f_{k-1}$ go
along the direction of $g_k$ to $\theta_1g_k$.
\begin{figure}[H]
\centering {\includegraphics[height=4cm,width=5cm]{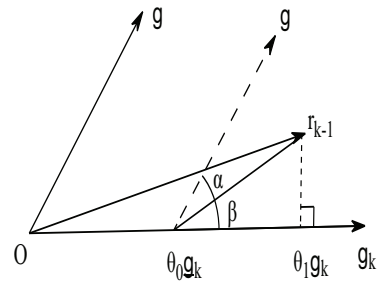}}
\caption{The drawback of boosting  }
\end{figure}

Under this circumstance, an advisable method is to control the
step-size in the linear search step  of Algorithm 1. Thus, various
variants of boosting, comprising the RTboosting, RSboosting and
$\varepsilon$-boosting, have been developed based on  different
strategies to control the step-size.   It is obvious that  the main
difficulty of these schemes  roots in how to select an appropriate
step-size. If the step size is too large, then these algorithms may
face the same problem as that of Algorithm \ref{alg1}. On the other
hand, if the step size is too small, then the numerical convergence
rate is also fairly slow \cite{Buhlmann2007}.

Different from the aforementioned strategies that focus on
controlling  the step-size of   $g_k^*$, we drive a novel direction
to improve the numerical convergence rate and consequently, the
generalization capability of boosting. The core idea is that if the
approximation (or learning) effect of the  $k$-th iteration is not
good, then we regard  $f_k$ to be too aggressive and therefore
  shrink it within a certain extent. That is, if a new iteration
is employed, then we   impose  a re-scale operator on the
  estimator $f_k$. This is the reason why we call our new
strategy as the re-scale boosting (RBoosting). The following
Algorithm 2 depicts the main idea of RBoosting.

\begin{algorithm}[H]\caption{Re-scale boosting}\label{alg2}
\begin{algorithmic}
\STATE {{ Step 1 (Initialization)}: Given data
$\{(X_i,Y_i):i=1,\dots,m\}$, weak learner set $S$, a set of
shrinkage degree $\{\alpha_k\}_{k=1}^\infty$,  iteration number
$k^*$, and $f_0\in\mbox{Span}(S)$}.
 \STATE{ { Step 2 (Projection of gradient)}: Find $g_k^*\in S$ such that
$$
              -Q_m'(f_{k-1},g_k^*)=\sup_{g\in S}
              - Q_m'(f_{k-1},g).
$$}
 \STATE{{Step 3 (Linear search)}:
 Find $\beta_k^*\in\mathbf R$  such that
$$
      Q_m((1-\alpha_k)f_k+\beta_k^*g^*_k)= \inf_{\beta_k\in\mathbf R}Q_m((1-\alpha_k)f_k+\beta_kg^*_k)
$$
Update $f_{k+1}=(1-\alpha_k)f_k+\beta_k^*g_k^*$.}
 \STATE{ { Step 4
(Iteration)}: Increase $k$ by one and repeat Step 2 and Step 3 if
$k< k^*$.}
\end{algorithmic}
\end{algorithm}

Compared Algorithm \ref{alg2} with Algorithm \ref{alg1}, the only
difference is that we employ  a re-scale operator $(1-\alpha_k)f_k$
in the linear search step of RBoosting. Here and hereafter, we call
$\alpha_k$ as the shrinkage degree. It can be easily found that
RBoosting is similar as the   greedy algorithm with free relaxation
(GAFR) \cite{Temlyakov2012} and   the $X$-greedy algorithm with
relaxation (XGAR)\footnote{In \cite{Zhang2003}, XGAR was called as
the sequential greedy algorithm, while in \cite{Barron2008}, XGAR
was named as the relaxed greedy algorithm for brevity.}
\cite{Temlyakov2008,Zhang2003} in sparse approximation. In fact,
RBoosting can be regarded as a marriage  of GAFR  and
 XGAR. To be detailed,  we adopt the
projection of gradient   of GAFR and  the linear search of XGAR to
develop Algorithm \ref{alg2}.

It should be also pointed out that the present paper is not the
first one to apply   relaxed greedy-type algorithms in the realm of
boosting. In particular, for the $L_2$ loss, XGAR has already been
utilized to design a boosting-type algorithm for regression in
\cite{Bagirov2010}. Since
  in both GAFR and XGAR, one needs to
tune two parameters simultaneously  in
 an optimization problem, GAFR and XGAR are time-consuming when faced with a general
convex loss function. This problem   is successfully avoided in
RBoosting.

\section{Theoretical behaviors of RBoosting}
In this section, we study the theoretical behaviors  of RBoosting.
We hope to address three basic issues regarding RBoosting, including
its numerical convergence rate,   consistency and generalization
error estimate.

To state the main results, some assumptions concerning the loss
function $\phi$ and dictionary $S$ should be presented. The first
one is a boundedness assumption of   $S$.

\begin{assumption}\label{ASSUMPTION 1}
 For arbitrary $g\in S$ and $x\in\mathcal X$, there exists a
constant $C_1$ such that
$$
       \sum_{i=1}^ng_i^2(x) \leq C_1.
$$
\end{assumption}

Assumption \ref{ASSUMPTION 1} is certainly a bit stricter than the
assumption
 $\sup_{g\in S,x\in\mathcal
X}|g_i(x)|\leq 1$ in \cite{Temlyakov2012,Zhang2005}. Introducing
such a condition is only for the purpose of deriving a fast
numerical convergence rate of RBoosting with general convex loss
functions. In fact, for a concrete loss function such as the $L_p$
loss with $1\leq p\leq\infty$, Assumption \ref{ASSUMPTION 1} can be
relaxed to $\sup_{g\in S,x\in\mathcal X}|g_i(x)|\leq 1$
\cite{Temlyakov2008}. Assumption \ref{ASSUMPTION 1} essentially
depicts the localization properties of the weak learners. Indeed, it
states that, for arbitrary fixed $x\in\mathcal X$, expert for a
small number of weak
 learners, all the $|g_i(x)|'s$ are very small. Thus, it   holds
 for  almost all the  widely
used weak learns   such as the  trees \cite{Friendman2001}, stumps
\cite{Zhang2005}, neural networks \cite{Bagirov2010} and splines
\cite{Buhlmann2003}. Moreover, for arbitrary dictionary
$S'=\{g_1',\dots,g_n'\}$, we can rebuild it as $
S=\{g_1,\dots,g_n\}$ with
$g_i=g_i'/(\sqrt{\sum_{i=1}^n(g_i')^2(x)})$.  It should be noted
that Assumption 1 is the only condition concerning the dictionary
throughout the paper, which is different from
\cite{Bartlett2007,Zhang2005} that additionally imposed either
VC-dimension or Rademacher complexity constraints to the weak
learner set $S$.

We then give some restrictions to the loss function,  which have
already adopted in
\cite{Bartlett2007,Bickel2006,Zhang2003,Zhang2005}.

\begin{assumption}\label{ASSUMPTION 2}
 (i) If $|f(x)|\leq \mathcal R_1$, $|y|\leq \mathcal R_2$, then
there
 exists a  continuous function $H_\phi$ such that
\begin{equation}\label{bound for loss}
         |\phi(f,y)|\leq H_\phi(\mathcal R_1,\mathcal R_2).
\end{equation}

(ii) Let    $\mathcal D=\{f:Q_m(f)\leq Q_m(0)\}$ and $f^*=\min_{f\in
\mathcal D}Q_m(f)$. Assume that $\forall c_1,c_2$ satisfying
$Q_m(f^*)\leq c_1<c_2\leq Q_m(0)$, there holds
\begin{eqnarray}\label{Assumption to loss1}
      &0&
      \leq \inf\{Q^{''}_m(f,g):c_1<Q(f)<c_2,g\in
      S\}\\
      &\leq&
      \sup\{Q^{''}_m(f,g):Q_m(f)<c_2, h\in S\}<\infty. \label{Assumption to loss2}
\end{eqnarray}
\end{assumption}

It should be pointed out that (i) concerns the boundedness of $\phi$
and therefore is mild. In fact, if $\mathcal R_1$ and $\mathcal R_2$
are bounded, then (i) implies that $\phi(f,y)$ is also bounded. It
is obvious that (i) holds for almost all commonly  used loss
functions. Once $\phi$ is given, $H_\phi(\mathcal R_1,\mathcal R_2)$
can be determined directly. For example, if $\phi$ is the $L_2$ loss
for regression, then $H_\phi(\mathcal R_1,\mathcal R_2)\leq(\mathcal
R_1+\mathcal R_2)^2$; if $\phi$ is the exponential loss for
classification, then $\mathcal R_2=1$ and $H_\phi(\mathcal
R_1,\mathcal R_2)\leq \exp\{\mathcal R_1\}$; if $\phi$ is the
logistic loss for classification, then $H_\phi(\mathcal R_1,\mathcal
R_2)\leq
 \log (1+\exp\{\mathcal R_1\})$.

 As
$Q_m(f)=\sum_{i=1}^m\phi(f(X_i),Y_i)$,  conditions (\ref{Assumption
to loss1}) and (\ref{Assumption to loss2}) actually describe the
strict convexity and smoothness of $\phi$ as well as $Q_m$.
Condition (\ref{Assumption to loss1}) guarantees the strict
convexity of $Q_m$ in a certain direction. Under this condition, the
maximization (and  minimization) in   projection of gradient step
(and linear search step) of Algorithms \ref{alg1} and \ref{alg2} are
well defined. Condition (\ref{Assumption to loss2}) determines the
smoothness
  property of   $Q_m(f)$. For arbitrary
   $f(x)\in[-\lambda,\lambda]$, define the first and second moduli of smoothness
of  $Q_m(f)$   as
$$
        \rho_1(Q_m,u)=\sup_{f,\|h\|=1}|Q_m(f+uh)-Q_m(f)|,
$$
and
\begin{eqnarray*}
        \rho _2(Q_m,u)
        &=&  \sup_{f,\|h\|=1}|Q_m(f+uh)\\
        &+&
        Q_m(f-uh)-2Q_m(f)|,
\end{eqnarray*}
where $\|\cdot\|$ denotes the uniform norm. It is easy to deduce
that if (\ref{Assumption to loss2}) holds, then there exist
 constants $C_2$ and $C_3$ depending only on $\lambda$ and $c_2$ such that
\begin{equation}\label{Assumption to loss 3}
           \rho_1(Q_m,u)\leq C_2\|u\|,\ \mbox{and}\  \rho_2(Q_m,u)\leq C_3\|u\|^2.
\end{equation}
It is easy to verify that all the widely used loss functions such as
the $L_2$ loss, exponential loss and logistic loss satisfy
Assumption 2.

By the help of the above stations, we are in a position to present
the first theorem, which focuses on  the numerical convergence rate
of RBoosting.

\begin{theorem}\label{NUMERICAL CONVERGENCE}
Let $f_k$ be the estimator defined by Algorithm \ref{alg2}. If
Assumptions \ref{ASSUMPTION 1} and \ref{ASSUMPTION 2} hold and
$\alpha_k=\frac3{k+3}$, then  for any
  $h\in\mbox{Span}(S)$, there holds
\begin{equation}\label{numerical convergence}
       Q_m(f_k)-Q_m(h)\leq C(\|h\|_1^2+\log k)k^{-1},
\end{equation}
where $C$ is a constant depending only on $c_1,c_2,C_1$, and
$$
        \|h\|_1=\inf_{(a_j)_{j=1}^n\in\mathbf R^n}\sum_{j=1}^n|a_j|,\ \mbox{for}\
        h=\sum_{j=1}^na_jg_j.
$$
\end{theorem}

If $\phi(f,y)=(f(x)-y)^2$ and $S$ is an orthogonal basis, then there
exists an $h^*\in\mbox{Span} (S) $ with bounded $\|h^*\|_1$ such
that   \cite{DeVore1996}
$$
         |Q_m(f_k)-Q_m(h^*)|\geq C k^{-1},
$$
where $C$ is an absolute constant. Therefore, the numerical
convergence rate deduced in (\ref{numerical convergence}) is almost
optimal in the sense that for at least some loss functions (such as
the $L_2$ loss) and  certain dictionaries (such as the orthogonal
basis), up to a logarithmic factor, the deduced rate is optimal.
%It
%can be found in the proof of Theorem \ref{NUMERICAL CONVERGENCE}
%that the main reason for the occurrence of the logarithmic factor is
%that   the  boundedness of $|f_k(x)|$ for arbitrary loss functions
%is not confirmed.  In fact, we  only prove   $|f_k(x)|\leq C\log k$.
%If   additional constraints to the loss function, an extreme case of
%which  is the $l_2$ loss, are imposed,  then the boundedness of
%$f_k$ can be easily deduced \cite{Temlyakov2008}.
% Then the same method as in this paper
%can omit the logarithmic factor in (\ref{numerical convergence}).
%That is, we can actually   prove the optimality of the $l_2$ regret
%boosting. The numerical convergence rate (\ref{numerical
%convergence}) is only presented for any loss functions satisfying
%Assumption 2 rather than a concrete one.
Compared with the relaxed greedy algorithm for convex optimization
\cite{DeVore2014,Temlyakov2012} that  achieves the optimal numerical
convergence rate, the   rate derived in (\ref{numerical
convergence}) seems a bit slower. However, in
\cite{DeVore2014,Temlyakov2012}, the set $\mathcal D=\{f:Q_m(f)\leq
Q_m(0)\}$ is assumed to  be bounded. This is a quite strict
assumption  and, to the best of our knowledge, it is difficult to
verify whether the widely used $L_2$ loss, exponential loss and
logistic loss satisfy this condition. In Theorem \ref{NUMERICAL
CONVERGENCE}, we omit this condition in the cost of adding an
additional logarithmic factor to the numerical convergence rate and
some other easy-checked assumptions to the loss function and
dictionary.

Finally, we  give an explanation why we select the shrinkage degree
$\alpha_k$ as $\alpha_k=\frac{3}{k+3}$. From the definition of
$f_k$, it follows that the numerical convergence rate may depend on
the shrinkage degree. In particular, Bagirov et al.
\cite{Bagirov2010}, Barron et al. \cite{Barron2008} and Temlyakov
\cite{Temlyakov2008} used different $\alpha_k$ to derive the optimal
numerical convergence rates of   relaxed-type greedy algorithms.
After checking our proof, we find that our result remains correct
for arbitrary $\alpha_k=\frac{C_4}{C_5k+C_6}<1$ with $C_4,  C_5,
C_6$ some finite positive integers. The only   difference is that
the constant $C$ in (\ref{numerical convergence}) may be different
for different $\alpha_k$. We select $\alpha_k=\frac3{k+3}$ is only
for the sake of brevity.

Now we turn to derive both the consistency and learning rate of
RBoosting. The consistency of  the boosting-type algorithms
describes whether the
  risk of     boosting can approximate the Bayes risk
 within arbitrary accuracy when $m$ is large
enough, while the learning rate depicts its convergence rate.
Several authors have shown that Algorithm \ref{alg1} with some
specific loss functions is consistent. Three most important results
can be found in \cite{Bartlett2007,Bickel2006,Jiang2004}. Jiang
\cite{Jiang2004} proved a process consistency property for Algorithm
\ref{alg1}, under certain assumptions. Process consistency means
that there exists a sequence $\{t_m\}$ such that if boosting with
sample size $m$ is stopped after $t_m$ iterations, its risk
approaches the Bayes risk. However, Jiang   imposed strong
conditions on the underlying distribution: the distribution of $X$
has to be absolutely continuous with respect to the Lebesgue
measure. Furthermore, the result derived in \cite{Jiang2004} didn't
give any hint on when the algorithm should be stopped since the
proof was not constructive. \cite{Bartlett2007,Bickel2006}  improved
the result of \cite{Jiang2004} and demonstrated that a simple
stopping rule is sufficient for consistency: the number of
iterations is a fixed function of $m$. However, it can also be found
in \cite{Bartlett2007,Bickel2006} that the deduced learning rate was
fairly slow.  \cite[Th.6]{Bartlett2007} showed that the risk of
boosting converges to the Bayes risk within a logarithmic speed.

Without loss of generality, we assume $|Y_i|\leq M$ almost surely
with $M> 0$. The following Theorem \ref{LEARNING RATE BOOSTING}
plays a  crucial role in deducing both the consistency and fast
learning rate of RBoosting.

\begin{theorem}\label{LEARNING RATE BOOSTING}
Let $f_k$ be the estimator obtained in Algorithm \ref{alg2}. If
$\alpha_k=\frac3{k+3}$ and Assumptions \ref{ASSUMPTION 1} and
\ref{ASSUMPTION 2} hold, then for arbitrary $h\in\mbox{Span} (S),$
there holds
\begin{eqnarray*}
         \mathbf E\{  Q(f_k)-Q(h)\}
          \leq
          C(\|h\|_1^2+\log k)k^{-1} &&\\
         +  C'(H_\phi(\log k,M)+H_\phi(\|h\|_1,M))\frac{k(\log m+\log
                    k)}{m}, &&\nonumber
\end{eqnarray*}
where $C$ and $C'$ are constants depending only on $c_1,c_2$ and
$C_1$.
\end{theorem}

Before giving the consistency of RBoosting, we should give some
explanations and remarks to Theorem \ref{LEARNING RATE BOOSTING}.
Firstly, we present the values of $H_\phi(\log k,M)$ and
$H_\phi(\|h\|_1,M)$. Taking $H_\phi(\log k,M)$ for example,  if
$\phi$ is the $L_2$ loss for regression, then $H_\phi(\log
k,M)=(\log k+M)^2$, if $\phi$ is the logistic loss for
classification, then $H_\phi(\log k,M)=\log (k+1)$ and if $\phi$ is
the exponential loss for classification, then $H_\phi(\log k,M)=k$.
Secondly, we provide a simple method to improve the bound in Theorem
\ref{LEARNING RATE BOOSTING}. Let
$\pi_Mf(x):=\min\{M,|f(x)|\}\mbox{sgn}(f(x))$ be the truncation
operator at level $M$. As $Y\in[-M,M]$ almost surely, there holds
\cite{Zhou2006}
$$
            \mathbf E\{  Q(\pi_Mf_k)-Q(h)\}\leq \mathbf E\{
            Q(f_k)-Q(h)\}.
$$
Noting that there is not any computation to do such a truncation,
  this  truncation technique has been widely used to rebuild
the estimator and improve  the learning rate of boosting
\cite{Bagirov2010,Barron2008,Bartlett2007,Bickel2006}. However, this
 approach has a drawback: the usage of the truncation operator
entails that the estimator $\pi_Mf_k$ is (in general) not an element
of Span$(S)$. That is, one aims to build an estimator in   a class
 and actually obtains  an estimator out of it. This is the reason why
we do not introduce the truncation operator in Theorem \ref{LEARNING
RATE BOOSTING}. Indeed, if we use the truncation operator, then  the
same method as that in the proof of Theorem \ref{LEARNING RATE
BOOSTING} leads to the following Corollary \ref{LEARNING RATE
BOOSTING of TRUNCATION}.

\begin{corollary}\label{LEARNING RATE BOOSTING of TRUNCATION}
Let $f_k$ be the estimator obtained in Algorithm \ref{alg2}. If
$\alpha_k=\frac{3}{k+3}$ and Assumptions \ref{ASSUMPTION 1} and
\ref{ASSUMPTION 2} hold, then for arbitrary $h\in\mbox{Span}(S),$
there holds
\begin{eqnarray}
         &&\mathbf E\{  Q(\pi_Mf_k)-Q(h)\}
          \leq
          C(\|h\|_1^2+\log k)k^{-1}\nonumber \\
          &&+  C'(H_\phi(M,M)+H_\phi(\|h\|_1,M))\frac{k(\log m+\log
                    k)}{m},\nonumber
\end{eqnarray}
where $C$ and $C'$ are constants depending only on $c_1,c_2$ and
$C_1$.
\end{corollary}

 By the help of Theorem \ref{LEARNING RATE BOOSTING}, we can derive  the consistency of
 RBoosting.

\begin{corollary}\label{CONSISTENCY}
Let $f_k$ be the estimator obtained in Algorithm \ref{alg2}. If
$\alpha_k=\frac3{k+3}$, Assumptions \ref{ASSUMPTION 1} and
\ref{ASSUMPTION 2} hold and
\begin{equation}\label{selection of k}
        k\rightarrow\infty, \frac{H_\phi(\log k,M)k\log
        m}{m}\rightarrow 0, \mbox{when} \ m\rightarrow\infty,
\end{equation}
then
$$
          \mathbf E\{  Q(f_k)\}\rightarrow
          \inf_{f\in\mbox{Span}(S)}Q(f),\  \mbox{when} \ m\rightarrow\infty.
$$
\end{corollary}

Corollary \ref{CONSISTENCY} shows that if the number of iterations
satisfies (\ref{selection of k}), then RBoosting is consistent. We
should point out that if the loss function is specified, then, we
can deduce a concrete relation between $k$ and $m$ to yield the
consistency. For example, if $\phi$ is the logistic function, then
the condition (\ref{selection of k}) becomes $
    k\sim m^{\gamma}
$ with $0<\gamma<1$. This condition is somewhat looser than the
previous studies concerning the consistency of boosting
\cite{Bartlett2007,Bickel2006,Jiang2004} or its modified version
\cite{Barron2008,Zhang2005}.

When used to both classification and regression, there usually is an
overfitting resistance phenomenon  of boosting as well as its
modified versions \cite{Buhlmann2003,Zhang2005}. Our result shown in
Corollary \ref{CONSISTENCY} looks to contradict it at the first
glance, as $k$ must be smaller than $m$. We  illustrate that this is
not the case. It can be found in \cite{Buhlmann2003,Zhang2005} that
expect for Assumption \ref{ASSUMPTION 1}, there is another condition
such as the covering number, VC-dimension, or Rademacher complexity
imposed to the dictionary. We highlight  that if the dictionary of
RBoosting is endowed with a similar assumption, then   the condition
$k<m$ can be omitted  by using the similar methods   in
\cite{Bickel2006,Lin2013,Zhang2005}. In short, our assertions show
that whether RBoosting is overfiiting resistant depends on the
dictionary.

At last, we give a learning rate analysis of RBoosting, which is
also a consequence of Theorem \ref{LEARNING RATE BOOSTING}.

\begin{corollary}\label{RATE}
Let $f_k$ be the estimator obtained in Algorithm 2. Suppose that
$\alpha_k=\frac3{k+3}$  and Assumptions 1 and 2 hold. For arbitrary
$h\in\mbox{Span} (S),$ if $k$ satisfies
\begin{equation}\label{Select of k 22222}
       k\sim\sqrt{\frac{m}{H_\phi(\log k,M) +H_\phi(\|h\|_1,M)}},
\end{equation}
then there holds
\begin{eqnarray}\label{corollary3}
         &&\mathbf E\{  Q(f_k)-Q(h)\}\\
          &&\leq
            C'(\sqrt{H_\phi(\log m,M)+H_\phi(\|h\|_1,M)}\nonumber\\
            &&+\|h\|_1^2) m^{-1/2}\log m,\nonumber
\end{eqnarray}
where $C$ and $C'$ are constants depending only on $c_1,c_2$ and
$C_1$ and $M$.
\end{corollary}

The learning rate (\ref{corollary3}) together with the stopping
criteria (\ref{Select of k 22222}) depends heavily on $\phi$. If
$\phi$ is the logistic loss for classification, then $H_\phi(\log
m,M)= \log( m+1) $ and $H_\phi(\|h\|_1,M)=\log (\|h\|_1+1)$, we thus
derive from (\ref{corollary3}) that,
$$
        \mathbf E\{  Q(f_k)-Q(h)\}
         \leq
            C'(\log (m+1)+\|h\|_1^2) m^{-1/2}\log
            m.
$$
We encourage the readers to compare our result with
\cite[Th.3.2]{Zhang2005}.  Without the Rademacher assumptions,
RBoosting theoretically performs at least the same as that of
RTboosting. If $\phi$ is the $L_2$ loss for regression, we can
deduce that
$$
        \mathbf E\{  Q(f_k)-Q(h)\}
         \leq
            C'(\log m+\|h\|_1^2) m^{-1/2}\log
            m,
$$
which is almost the same as the result in \cite{Bagirov2010}. If
$\phi$ is the exponential loss for classification, by setting $k\sim
m^{1/3}$, we can derive
$$
        \mathbf E\{  Q(f_k)-Q(h)\}
         \leq
            C'(\log m+e^{\|h\|_1}) m^{-1/3}\log
            m,
$$
which is much faster than AdaBoost \cite{Bartlett2007}. It should be
noted that if the truncation operator is imposed to the RBoosting
estimator, then the learning rate of  the re-scale AdaBoost can also
be improved to
$$
        \mathbf E\{  Q(\pi_Mf_k)-Q(h)\}
         \leq
            C'(\log m+e^{\|h\|_1}) m^{-1/2}\log
            m.
$$

\section{Numerical Results}
In this section,  we conduct a series  of toy simulations   and
 real data  experiments to demonstrate  the promising
outperformance of the proposed RBoosting over the original boosting
algorithm. For comparison,  three other popular boosting-type
algorithms, i.e., $\epsilon$-boosting \cite{Hastie2007}, RSboosting
\cite{Friendman2001} and RTboosting
 \cite{Zhang2005}, are also considered.  In the following
experiments, we utilize the $L_2$ loss function for regression
(namely, L2Boost)  and logistic loss function for classification
(namely, LogitBoost).
 %It is known that the
%boosting trees algorithm also requires  the specification of the
%number of splits (or the number of nodes) that are used for fitting
%each regression or classification tree. The number of leaves equals
%the number of splits plus one. Specifying $J$ splits corresponds to
%an estimate with up to $J$-way interactions.
Furthermore, we use the CART \cite{Breiman1984} (with the number of
splits $J=4$) to build up the week learners for regression tasks in
the toy simulations
 and
decision stumps (with the number of splits $J=1$) to build up the
week learners for regression tasks in real data experiments and all
classification tasks.

\subsection{Toy simulations}
We first consider  numerical simulations for regression problems.The
data are drawn from the  following model:
\begin{equation}
Y = m(X)+\sigma\cdot\varepsilon,
\end{equation}
where  $X$ is uniformly distributed on $[-2, 2]^d$ with $d\in\{1,
10\}$, $\varepsilon$, independent of $X$, is  the standard gaussian
noise  and the noise level $\sigma$ varies among in $\{0, 0.3, 0.6,
1\}$.    Two typical regression functions \cite{Bagirov2010} are
considered in  the simulations.  One is a univariate piecewise
function defined by
\begin{equation}\label{regression1}
        m_1(x)= \left\{ \begin{array}{cc}
        10\sqrt { - x} \sin (8\pi x), & \leq x < 0, \\
               0,  &   \mbox{else},
\end{array} \right.
\end{equation}
and the other is a multivariate continuously differentiable sine
function defined as
\begin{equation}\label{regression2}
               m_2(x)= \sum\limits_{j = 1}^{10} {{{( - 1)}^{j - 1}}{x_j}\sin ({x_j}^2)}.
\end{equation}

For these  regression functions and all values of $\sigma$, we
generate  a training set of  size 500,  and then collect  an
independent validation data set of size 500 to select the parameters
of each boosting algorithms: the number of iterations $k$, the
regularization parameter $\nu$ of RSboosting, the truncation value
of RTboosting, the shrinkage degree of RBoosting and $\varepsilon$
of $\varepsilon$-boosting. In  all the  numerical examples, we chose
$\nu$ and  $\epsilon$   from a 20 points set whose elements are
uniformly localized in $[0.01,1]$. We select  the truncated value of
RTboosting   the same as that in \cite{Zhang2005}. To tune the
shrinkage degree, $\alpha_k=2/(k+u),$ we employ  20  values of $u$
which are drawn   logarithmic equally spaced between $1$ to $10^6$.
To compare the performances of all the mentioned  methods, a test
set of 1000 noiseless observations   is used to evaluate the
performance in terms of the root mean squared error (RMSE).

\begin{table*}[htb]
\begin{center}
 \caption{Performance comparison of different boosting algorithms on simulated regression data examples}\label{para}
 %\begin{minipage}{0.5\textwidth}
 \begin{tabular}{|c|c|c|c|c|c|}\hline
 $\sigma$&  Boosting  & RSboosting & RTboosting & RBoosting & $\epsilon$-boosting    \\ \hline
 \multicolumn{6}{|c|}{piecewise function (\ref{regression1})} \\ \hline
 %\multicolumn{8}{|c|}{$f_1$  } \\ \hline
 0    &0.2698(0.0495) &0.2517(0.0561) &0.3107(0.0905) &{0.2460(0.0605)} & {\bf 0.2306}(0.0827) \\ \hline
 0.3  &0.6204(0.0851) &{\bf 0.4635}(0.0728) &0.5131(0.0735) & {0.5112(0.0779)} & 0.4844(0.0862) \\ \hline
 0.6  &0.7339(0.0706) &0.7317(0.0392) &0.7475(0.0333) & {{\bf 0.7206}(0.0486)} & 0.7403(0.0766) \\ \hline
 1    &1.1823(0.0483) &1.1474(0.0485) &1.1776(0.0604) & {1.1489(0.0485)} & {\bf1.1395}( 0.0590) \\ \hline
 \multicolumn{6}{|c|}{continuously differentiable sine functions (\ref{regression2}) } \\ \hline
 %\multicolumn{8}{|c|}{$f_1$  } \\ \hline
 0    &2.3393(0.1112) &1.7460(0.0973) &1.8388(0.1102) & {{\bf 1.6166}(0.0955)} & 1.7434(0.0804) \\ \hline
 0.3  &2.4051(0.1112) &1.7970(0.0951) &1.8380(0.0830) & {{\bf 1.6732}(0.0928)} & 1.7665(0.0718) \\ \hline
 0.6  &2.4350(0.0836) &1.8866(0.0837) &1.9628(0.0853) & {{\bf 1.7730}(0.0832)} & 1.8895(0.0880) \\ \hline
 1    &2.6583(0.1103) &2.0671(0.0789) &2.1575(0.0891) & {{\bf 1.9870}(0.1092)} & 2.0766(0.0956) \\ \hline
  \end{tabular}
 %\end{minipage}
 \end{center}
 \end{table*}

Table I documents the mean RMSE over 50 independent runs.  The
standard errors are also reported (numbers in parentheses). Several
observations can be easily drawn from Table I. Firstly, concerning
the generalization capability, all the variants essentially
outperform the original boosting algorithm. This is not a surprising
result since all the variants introduce an additional parameter.
Secondly, RBoosting    performs as the almost   optimal variant
since its RMSEs are the smallest or  almost smallest for all the
simulations. This means that, if we only focus on the generalization
capability, then RBoosting is a preferable choice.

In the second  toy simulation, we consider  the ``orange data''
model which was used in \cite{Zhu2003} for binary classification. We
generate 100 data points for  each class to build up the training
set. Both  classes have two independent standard normal inputs
$x_1$, $x_2$, but the inputs for the second class conditioned on
$4.5\leq x_1^2+x_2^2\leq 8$. Similarly, to make the classification
more difficult, independent feature noise $q$ were added to
 the inputs.  One can find more details about
this data set in \cite{Zhu2003}.

\begin{table*}[htb]
\begin{center}
 \caption{Performance comparison of different boosting algorithms on simulated ``orange data"  example}\label{para}
 %\begin{minipage}{0.5\textwidth}
 \begin{tabular}{|c|c|c|c|c|c|}\hline
 $q$& Boosting       & RSboosting     & RTboosting     & RBoosting        & $\epsilon$-boosting    \\ \hline
 %\multicolumn{8}{|c|}{$f_1$  } \\ \hline
 0  &11.19(1.32)\% &10.36(1.16)\% &10.50(1.19)\% &{{ 10.44}(1.12)\%} & {\bf 10.29}(1.17)\% \\ \hline
 2  &11.27(1.29)\% &{\bf 10.48}(1.24)\% &10.71(1.19)\% &{{10.59}(1.25)\%} & 10.60(1.28)\% \\ \hline
 4  &11.79(1.54)\% &{\bf 10.79}(1.21)\% &11.07(1.41)\% &{{10.90}(1.24)\%} & 10.94(1.26)\% \\ \hline
 6  &12.02(1.62)\%  &{ 10.93}(1.21)\%  &{ 11.20}(1.23)\%  &{\bf 10.91}(1.28)\% & {11.02}(1.32)\%  \\ \hline
 %8  &6.79(0.0870)\%  &6.88(0.0772)\%  &14.70(0.0824)\%  &{6.75(0.0838)\%}  & {\bf 6.52}(0.0887)\%  \\ \hline
 \end{tabular}
 %\end{minipage}
 \end{center}
 \end{table*}

Table II reports  the classification accuracy of five boosting-type
algorithms over 50 independent runs.  Numbers in parentheses are the
standard errors. In this simulation, for  $q$ varies among   $\{0,
2, 4, 6\}$, we generate  a validation set of size 200 for tuning the
parameters, and then 4000 observations
 to evaluate the performances in terms of classification error.  For
this classification task, RBoosting outperforms the original
boosting in terms of the generalization error. It can also be found
that as far as the classification problem is concerned, RBoosting is
at least comparable to  other variants.
  Here we do not compare the performance
with the performance of SVMs reported in \cite{Zhu2003}, because the
main purpose of our simulation is to highlight the outperformance of
the proposed RBoosting  over the original boosting.

All the above toy simulations from   regression to classification
verify the theoretical assertions in the last section and illustrate
the merits of RBoosting.

\subsection{Real Data Examples}
In this subsection,  We  pursue the  performance of RBoosting   on
eight real data sets (the first five data sets for regression and
the others for classification).

The first  data set is the Diabetes data set\cite{Efron2004}. This
data set contains 442 diabetes patients that are measured on ten
independent variables, i.e., age, sex, body mass index etc. and one
response variable, i.e., a measure of disease progression. The
second  one is the Boston Housing data set created from a housing
values survey in suburbs of Boston by Harrison and Rubinfeld
\cite{Harrison1978}. This data set contains 506 instances which
include thirteen attributions, i.e., per capita crime rate by town,
proportion of non-retail business acres per town, average number of
rooms per dwelling etc. and one response variable, i.e., median
value of owner-occupied homes. The third one is the Concrete
Compressive Strength (CCS) data set created from \cite{Ye1998}.  The
data set contains 1030 instances including eight quantitative
independent variables, i.e., age and ingredients etc. and one
dependent variable, i.e., quantitative concrete compressive
strength. The fourth  one is  the Prostate cancer data set derived
from a study of prostate cancer by Blake et al. \cite{Blake1998}.
The data set consists of the medical records of 97 patients who were
about to receive a radical prostatectomy. The predictors are eight
clinical measures, i.e., cancer volume, prostate weight, age etc.
and one response variable, i.e., the logarithm of prostate-specific
antigen. The fifth one is the Abalone data set, which comes from an
original study in \cite{Nash1994} for predicting the age of abalone
from physical measurements. The data set  contains 4177  instances
which were measured on eight independent variables, i.e., length,
sex, height etc. and one response variable, i.e., the number of
rings. For  classification task, three benchmark data sets are
considered, namely Spam, Ionosphere and WDBC, which can be obtained
from UCI Machine Learning Repository.  Spam data contains 4601
instances, and 57 attributes. These data are used to measure whether
an instance is considered to be spam. WDBC (Wisconsin Diagnostic
Breast Cancer) data contains 569 instances, and 30 features. These
data are used to identify whether an instance is diagnosed to be
malignant or benign. Ionosphere data contains 351 instances, and 34
attributes. These data are used to measure whether an instance was
``good" or ``bad".

\begin{table*}[htb] \addtolength{\tabcolsep}{-1pt}
\begin{center}
 \caption{Performance comparison of different boosting algorithms on real   data examples}\label{para}
 %\begin{minipage}{0.5\textwidth}
 \begin{tabular}{|c|c|c|c|c|c|}\hline
 dataset  & Boosting  & RSboosting  &  RTboosting   & RBoosting & $\epsilon$-boosting   \\ \hline
 %\multicolumn{8}{|c|}{$f_1$  } \\ \hline
 Diabetes &59.0371(4.1959)    &{\bf 55.3109}(3.6591)      &56.1343(3.2543)        &  55.6552(4.5351)   & 57.7947(3.3970)  \\ \hline
 Housing  &4.4126(0.5311)     &4.2742(0.7026)       &4.3685(0.3929)         &  4.1752(0.3406)    & {\bf 4.1244}(0.3322)    \\ \hline
 CCS      &5.4345(0.5473)     &{\bf 5.2049}(0.1678)       &5.5826(0.1901)         &  5.3711(0.1807)    & 5.9621(0.1960)   \\ \hline
 Prostate &0.3131(0.0598)     &0.1544(0.0672)       &0.2450(0.0631)         &  {\bf 0.1193}(0.0360)    & 0.1939(0.0545)   \\ \hline
 Abalone  &2.2180(0.0710)     &2.1934(0.0504)       &2.3633(0.0762)         &  {\bf 2.1922}(0.0574)    & 2.2098(0.0474)   \\ \hline
 Spam             & 6.06(0.60)\%    & 5.13(0.52)\%      & 5.24(0.48)\%       & 5.06(0.55)\%    & {\bf 5.02}(0.51)\% \\ \hline
 Ionosphere     & 8.27(2.88)\%    & 5.80(1.92)\%      & 6.09(2.24)\%       &{\bf 5.23}(2.31)\% & 5.92(2.64)\% \\ \hline
 WDBC          & 5.31(2.11)\%    &2.45(1.39)\% & { 2.69(1.58)}\%       & {\bf 2.09}(1.55)\%    & 2.52(1.33)\% \\ \hline
  \end{tabular}
 %\end{minipage}
 \end{center}
 \end{table*}

For each real data,  we  randomly (according to the uniform
distribution) select $50\%$ data for training, $25\%$ data to build
the validation set for tuning the parameters and the remainder
$25\%$ data as the test set for evaluating  the performances of
different boosting-type algorithms. We repeat  such randomization 20
times and report  the average errors and standard errors (numbers in
parentheses) in Table III. The parameter selection strategies of all
boosting-type algorithms are the same as those in the toy
simulations.  It   can be easily observed that, all the variants
outperform the original boosting algorithm to a large extent.
Furthermore, RBoosting at least performs as the second best
algorithm among all the variants. Thus, the results of real data
coincide with the toy simulations and therefore, experimentally
verify our theoretical assertions. That is,  all the experimental
results show that the new idea ``re-scale'' of RBoosting is
numerically efficient and comparable to the idea ``regularization''
of other variants of boosting. This paves a new road to improve the
performance of boosting.

\section{Proof of Theorem \ref{NUMERICAL CONVERGENCE}} To prove
Theorem \ref{NUMERICAL CONVERGENCE}, we need the following three
lemmas. The first one is a small generalization of \cite[Lemma
2.3]{Temlyakov2008}. For the sake of completeness, we give a simple
proof.

\begin{lemma}\label{NUMBER THEORY}
Let $j_0>2$ be a natural number. Suppose that three positive numbers
$c_1<c_2\leq j_0$, $\mathcal C_0$ be given. Assume that a sequence
$\{a_n\}_{n=1}^\infty$ has the following two properties:

(i) For all $1\leq n\leq j_0$,
$$     a_n\leq \mathcal C_0 n^{-c_1},
$$
 and,
for all $n\geq j_0$,
$$
        a_n\leq a_{n-1}+\mathcal C_0(n-1)^{-c_1}.
$$

(ii) If for some $v\geq j_0$ we have
$$
             a_v\geq \mathcal C_0v^{-c_1},
$$
then
$$
            a_{v+1}\leq a_v(1-c_2/v).
$$

Then, for all $n=1,2,\dots,$ we have
$$
            a_n\leq   2^{1+\frac{c_1^2+c_1}{c_2-c_1}} \mathcal
            C_0n^{-c_1}.
$$
\end{lemma}

\begin{IEEEproof}
For $1\leq v\leq j_0$, the inequality
$$
             a_v\geq \mathcal C_0v^{-c_1}
$$
implies that the set
$$
      V=\{v:a_v\geq \mathcal C_0v^{-c_1}\}
$$
does not contain $v=1,2,\dots,j_0$. We now prove that for any
segment $[n,n+k]\subset V$, there holds
$$
              k\leq (2^{\frac{c_1+1}{c_2-c_1}}-1)n.
$$
Indeed, let $n\geq j_0+1$ be such that $n-1\notin V$, which means
$$
          a_{n+j}\geq \mathcal C_0(n+j)^{-c_1},\ j=0,1,\dots,k.
$$
Then by the conditions (i) and (ii), we get
\begin{eqnarray*}
           a_{n+k}
           &\leq&
           a_n\Pi_{v=n}^{n+k-1}(1-c_2/v)\\
           &\leq&
           (a_{n-1}+\mathcal C_0(n-1)^{-c_1})\Pi_{n=n}^{n+k-1}(1-c_2/v).
\end{eqnarray*}
Thus, we have
$$
        (n+k)^{-c_1}\leq\frac{a_{n+k}}{\mathcal C_0}\leq2(n-1)^{-c_1}\Pi_{v=n}^{n+k-1}(1-c_2/v),
$$
where $c_2\leq j_0\leq v$. Taking logarithms and using the
inequalities
$$
      \ln (1-t)\leq -t, \ t\in[0,1);
$$
$$
      \sum_{v=n}^{m-1}v^{-1}\geq\int_{n}^m t^{-1}dt=\ln (m/n),
$$
we can derive that
\begin{eqnarray*}
      &&-c_1\ln\frac{n+k}{n-1}
      \leq
      \ln
      2+\sum_{v=n}^{n+k-1}\ln(1-c_2/v)\\
      &\leq&
      \ln2-\sum_{v=n}^{n+k-1}c_2/v\leq \ln2-c_2\ln\frac{n+k}{n}.
\end{eqnarray*}
Hence,
$$
        (c_2-c_1)\ln(n+k)\leq \ln2+(c_2-c_1)\ln
        n+c_1\ln\frac{n}{n-1},
$$
which implies
$$
           n+k\leq 2^{(c_1+1)/(c_2-c_1)}n
$$
and
$$
        k\leq  \left(2^{\frac{c_1+1}{c_2-c_1}}-1\right) n.
$$
Let us take any $m\in\mathbf N$. If $m\notin V$, we have the desired
inequality. Assume $m\in V$ and let $[n,n+k]$ be the maximal segment
in $V$ containing $m$, then we obtain
\begin{eqnarray*}
        a_m
        &\leq&
         a_n
         \leq a_{n-1}+\mathcal C_0(n-1)^{-c_1}\leq
        2\mathcal C_0(n-1)^{-c_1}\\
        &\leq&
        2\mathcal C_0m^{-c_1}\left(\frac{n-1}{m}\right)^{-c_1}.
\end{eqnarray*}
Since $k\leq \left(2^{\frac{c_1+1}{c_2-c_1}}-1\right) n$, we then
have
$$
       \frac{m}{n-1}\leq\frac{n+k}{n}\leq 2^\frac{c_1^2+c_1}{c_2-c_1}.
$$
This means that
$$
       a_m\leq 2\mathcal C_0m^{-c_1}2^\frac{c_1^2+c_1}{c_2-c_1},
$$
which finishes the proof of Lemma \ref{NUMBER THEORY}.
\end{IEEEproof}

 The convexity
of $Q_m$ implies that for any $f,g$,
$$
        Q_m(g)\geq Q_m(f)+ Q_m'(f,g-f),
$$
or, in other words,
$$
          Q_m(f)-Q_m(g)\leq  Q_m'(f,f-g)=- Q_m'(f,g-f).
$$
Based on this, we can obtain the following lemma, which was proved
in \cite[Lemma 1.1]{Temlyakov2012}.

\begin{lemma}\label{MODULUS OF SMOOTHNESS}
Let $Q_m$ be a Fr\'{e}chet differential convex function. Then the
following inequality holds for $f\in D$
$$
             0\leq Q_m(f+ug)-Q_m(f)-uQ_m'(f,g)\leq
             2\rho(A,u\|g\|).
$$
\end{lemma}

To aid the proof, we also need the following lemma, which can be
found in \cite[Lemma 2.2]{Temlyakov2001}.

\begin{lemma}\label{EQUALIVALENT MIN}
For any bounded linear $F$ and any dictionary $S$, we have
$$
       \sup_{g\in S}F(g)=\sup_{f\in \mathcal M_1(S)}F(f),
$$
where $\mathcal M_1(S)=\{\mbox{span}(S):\|f\|_1\leq 1\}$.
\end{lemma}

\begin{IEEEproof}[Proof of Theorem \ref{NUMERICAL CONVERGENCE}]
 We divide the proof into two steps. The
first step is to deduce an upper bound of $f_k$ in the uniform
metric. Since $f_{k+1}=(1-\alpha_{k+1})f_k+\beta_{k+1}^*g_{k+1}^*$,
we have
$$
          f_k=f_{k+1}+\frac{\alpha_{k+1}f_{k+1}-\beta_{k+1}^*g_{k+1}^*}{1-\alpha_{k+1}}.
$$
Noting $Q_m(f)$ is twice differential, if we use the Taylor
expansion around $f_{k+1}$, then
\begin{eqnarray*}
        Q_m(f_k)
        &=&
        Q_m(f_{k+1})\\
        &+&
        Q'_m\left(f_{k+1},\frac{\alpha_{k+1}f_{k+1}-\beta_{k+1}^*g_{k+1}^*}{1-\alpha_{k+1}}\right)\\
        &+&
        \frac12Q_m^{''}\left(\hat{f_k},\frac{\alpha_{k+1}f_{k+1}-\beta^*_{k+1}*g_{k+1}^*}{1-\alpha_{k+1}}\right)\\
        &=&
        Q_m(f_{k+1})+\frac{\alpha_{k+1}}{1-\alpha_{k+1}}Q'_m\left(f_{k+1},f_{k+1}\right)\\
        &-&
        \frac{\beta^*_{k+1}}{1-\alpha_{k+1}}Q'_m\left(f_{k+1},g^*_{k+1}\right)\\
        &+&
        \frac{\alpha^2_{k+1}}{2(1-\alpha_{k+1})^2}Q_m^{''}\left(\hat{f_k},f_{k+1}\right)\\
        &+&
        \frac{(\beta_{k+1}^*)^2}{2(1-\alpha_{k+1})^2}Q_m^{''}\left(\hat{f_k},g_{k+1}\right),
\end{eqnarray*}
where
$$
           \hat{f}=(1-\theta)\frac{\alpha_{k+1}f_{k+1}-\beta_{k+1}^*g_{k+1}^*}{1-\alpha_{k+1}}+\theta
f_{k+1}
$$
 for some $\theta\in (0,1)$. For the convexity of $Q_m$, we have
$$
     \frac{\alpha^2_{k+1}}{2(1-\alpha_{k+1})^2}Q_m^{''}\left(\hat{f_k},f_{k+1}\right)\geq
     0.
$$
Furthermore, if we use the fact that $f_{k+1}$ is the minimum on the
path from $(1-\alpha_{k+1}) f_k$ along $g^*_{k+1}$, then it is easy
to see that
$$
        Q'_m\left(f_{k+1},g^*_{k+1}\right)=0.
$$
According to the convexity of $Q_m$ again, we obtain
$$
       Q_m'(f_{k+1},f_{k+1})\geq Q_m(f_{k+1})-Q_m(0).
$$
Noting that $\frac{\alpha_{k+1}}{1-\alpha_{k+1}}=\frac4k$, we obtain
\begin{eqnarray*}
        Q_m(f_k)
        &\geq&
        Q_m(f_{k+1})
        +
        \frac4k(Q_m(f_{k+1})-Q_m(0))\\
        &+&
        \frac{(\beta_{k+1}^*)^2}{2}Q_m^{''}\left(\hat{f_k},g_{k+1}\right).
\end{eqnarray*}
If we write $\mathcal B=\inf\{Q_m''(f,g): c_1<Q_m(f)<c_2,g\in S\}$,
then we have
$$
       (\beta^*_{j+1})^2\leq\frac2{\mathcal
       B}\left(Q_m(f_j)-Q_m(f_{j+1})+\frac4jQ_m(0)\right).
$$
Therefore,
$$
         \sum_{j=0}^k(\beta_j^*)^2\leq \frac{2\log k}{\mathcal B}.
$$
Then it follows from the definition of $f_k$ that
\begin{eqnarray*}
           f_k
          &=&
          (1-\alpha_k)(1-\alpha_{k-1})\cdots(1-\alpha_2)\beta_1^*g_1^*\\
          &+&
          (1-\alpha_k)(1-\alpha_{k-1})\cdots(1-\alpha_3)\beta_2^*g_2^*\\
          &+&\dots
          +
          (1-\alpha_k)\beta^*_{k-1}g^*_{k-1}+\beta^*_kg_k^*.
\end{eqnarray*}
Therefore, it follows from the Assumption \ref{ASSUMPTION 1} that
\begin{equation}\label{boundedness of fk}
    |f_k(x)|\leq\sqrt{\sum_{j=0}^k(\beta_j^*)^2 \sum_{j=0}^n|g_j^*(x)|^2}
    \leq \sqrt{2C_1\log k/\mathcal B}.
\end{equation}

Now we turn to the second step, which derives the numerical
convergence rate of RBoosting. For arbitrary $\beta_k\in\mathbf R$
and $g_k\in S$, it follows form Lemma \ref{MODULUS OF SMOOTHNESS}
that
\begin{eqnarray*}
                   & & Q_m((1-\alpha_{k+1})f_{k}+\beta_{k+1}g_{k+1})\\
                     &=&
                    Q_m(f_{k}-\alpha_{k+1}f_{k}+\beta_{k+1}g_{k+1})\\
                    &\leq&
                    Q_m(f_{k})-\beta_{k+1}(-Q_m'(f_{k},g_{k+1}))\\
                    &-&
                    \alpha_{k+1}Q_m'(f_{k},f_{k})\\
                    &+&
                    2\rho(Q_m,\|\beta_{k+1}g_{k+1}-\alpha_{k+1}f_{k}\|).
\end{eqnarray*}
From Step 2 in Algorithm \ref{alg2},  $g^*_{k+1}$ satisfies
$$
            -Q_m'(f_{k},g^*_{k+1})=\sup_{g\in S}
          -Q_m'(f_{k},g).
$$
Set $\beta_k=\|h\|_1\alpha_k$.   It follows from Lemma
\ref{EQUALIVALENT MIN} that
\begin{eqnarray*}
          \sup_{g\in S}-Q_m'(f_{k},g)
         &=&
         \sup_{\phi\in
         \mathcal M_1(S)}-Q_m'(f_{k} ,\phi)\\
         &\geq&
         -\|h\|_1^{-1}   Q_m'(f_{k},h).
\end{eqnarray*}
  Under this circumstance, we get
\begin{eqnarray*}
             &&Q_m((1-\alpha_{k+1})f_{k}+\beta_{k+1}g^*_{k+1})\\
             &&\leq
             Q_m(f_{k})-\alpha_{k+1}
             (-Q_m'(f_{k},h-f_{k}))\\
             &+&
             2\rho(Q_m,\|\beta_{k+1}g^*_{k+1}-\alpha_{k+1}f_{k}\|).
\end{eqnarray*}
Based on Lemma \ref{MODULUS OF SMOOTHNESS}, we obtain
$$
          -Q_m'(f_{k} ,h-f_{k})\geq Q_m(f_{k})-Q_m(h).
$$
Thus,
\begin{eqnarray*}
          Q_m(f_{k+1})
          &=&
           Q_m((1-\alpha_{k+1})f_{k}+\beta_{k+1}g^*_{k+1})\\
           &\leq&
          Q_m(f_{k})-\alpha_k(Q_m(f_{k})-Q-m(h))\\
          &+&
          2\rho(Q_m,\left\|\|h\|_1\alpha_{k+1}g^*_{k+1}-\alpha_{k+1}f_{k}\right\|).
\end{eqnarray*}
Furthermore, according to (\ref{boundedness of fk}), we obtain
\begin{eqnarray*}
        &&\left\|\|h\|_1\alpha_{k+1}g^*_{k+1}-\alpha_{k+1}f_{k}\right\|\\
        &&\leq
        \|h\|_1\alpha_{k+1}+\alpha_{k+1}\|f_{k}\|\\
        &&\leq
        \|h\|_1\alpha_{k+1}+\alpha_{k+1}\|f_{k}\|_1\\
        &&\leq
        (\|h\|_1+\sqrt{2C_1\ln k})\alpha_{k+1}.
\end{eqnarray*}
Therefore,
\begin{eqnarray}
        &&Q_m(f_{k+1})-Q_m(h)
         \leq
        Q_m(f_{k})-Q_m(h) \nonumber\\
        &&-\alpha_{k+1}(Q_m(f_{k})-Q_m(h))  \nonumber \\
        &&+
        2\rho\left(Q_m,(\|h\|_1+\sqrt{2C_1\log
        k}/\mathcal B)\alpha_{k+1}\right). \label{important estimate}
\end{eqnarray}
Now, we use the above inequality and Lemma \ref{NUMBER THEORY} to
prove Theorem \ref{NUMERICAL CONVERGENCE}. Let
$a_k=Q_m(f_{k+1})-Q_m(h)$. Let $c_3\in (1,2]$ and $\mathcal C_0$ be
selected later. We then prove the conditions (i) and (ii) of Lemma
\ref{NUMBER THEORY} hold for an appropriately selected $\mathcal
C_0$. Set
$$
           \mathcal C_0=  2+\frac{A(0)}2+ \frac{72C_2}{25}\|h\|_1^2
           + \frac{288C_1C_2\log k}{25\mathcal B^2} .
 $$
  Then, it follows from (\ref{important estimate})
  and $\rho(Q_m,u)\leq C_2 u^2$ that
$$
            a_1\leq \frac{A(0)}4+ \frac{9C_2}{8}\|h\|_1^2\leq
            \mathcal C_0,\ a_2\leq \mathcal C_0 2^{-1},
$$
and for $v\geq 2$, there holds
$$
       a_v\leq a_{v-1}+\mathcal C_0(v-1)^{-1}.
$$
Thus the condition (i) of Lemma \ref{NUMBER THEORY} holds with
$j_0=2$.
 and $a_v\geq \mathcal C_0v^{-1}$, then by (\ref{important estimate}), we get
for $v\geq 6$,
\begin{eqnarray*}
     &&a_{v+1}
      \leq
      a_v(1-\alpha_{v+1}\\
      &&+
      2C_2  (\|h\|_1+\sqrt{2C_1\log
        k}/\mathcal B)^2\alpha_{v+1}^2/a_v)\\
      &&\leq
     a_v\left(1-\frac{3}{v+3}+\frac{1}{2v+2}\right)\\
     &&\leq
     a_v\left(1-\frac3{2v}\right).
\end{eqnarray*}
Thus the condition (ii) of Lemma \ref{NUMBER THEORY} holds with
$c_2=\frac32$. Applying Lemma \ref{NUMBER THEORY} we obtain
$$
       Q_m(f_k)-Q_m(h)\leq C(\|h\|^2_1+\log k)k^{-1},
$$
where $C$ is a constant depending only on $\mathcal B$, $C_1$ and
$C_2$.
 This finishes
the proof of Theorem \ref{NUMERICAL CONVERGENCE}.
\end{IEEEproof}

\section{Proof of Theorem \ref{LEARNING RATE BOOSTING}}

To aid the proof of Theorem \ref{LEARNING RATE BOOSTING}, we need
the following two technical lemmas, both of them can be found in
\cite{Zhou2006}.

Let $R>0$, we denote $  B_R$ as the closed ball of
$V_k=$Span$\{g_1^*,\dots,g_k^*\}$ with radius $R$ centered at
origin:
$$
              B_R=\{f\in V_k:\|f\|\leq R\}.
$$

\begin{lemma}\label{COVERING ESTIMATE}
For $R>0$ and $\eta>0$, we have
$$
            \log\mathcal N( B_R,\eta)\leq
            C_3k\log\left(\frac{4R}{\eta}\right),
$$
where $\mathcal N(   B_R,\eta)$ denotes the covering number of $
B_R$ with radius $\eta$ under the uniform norm.
\end{lemma}

The following ratio probability inequality is a standard result in
learning theory (see \cite{Zhou2006}).

\begin{lemma}\label{CONCENTRATION INEQUALITY}
Let $\mathcal G$ be a set of functions on $Z$ such that, for some
$c\geq 0$, $|g-\mathbf E(g)|\leq B$ almost everywhere and $\mathbf
E(g^2)\leq c\mathbf E(g)$ for each $g\in\mathcal G$. Then, for every
$\varepsilon>0$, there holds
\begin{eqnarray*}
             &&
             \mathbf P\left\{\sup_{f\in\mathcal G}
             \frac{\mathbf E(g)-\frac1m\sum_{i=1}^mg(z_i)}{\sqrt{
             \mathbf E(g)+\varepsilon}}\geq\sqrt{\varepsilon}\right\}\\
             &&\leq
             \mathcal N(\mathcal
             G,\varepsilon)\exp\left\{-\frac{m\varepsilon}{2c+\frac{2B}3}\right\}.
\end{eqnarray*}
\end{lemma}

\begin{IEEEproof}[Proof of Theorem \ref{LEARNING RATE BOOSTING}]
At first, we use the concentration inequality in Lemma
\ref{CONCENTRATION INEQUALITY} to bound
$$
           Q(f_k)-Q(h)
              -(Q_m(f_k)-Q_m(h)).
$$
We need apply Lemma
 \ref{CONCENTRATION INEQUALITY} to the set of functions
$\mathcal F_R$,
 where
$$
        \mathcal F_{R}:=\left\{\psi(Z)=\phi(f(X),Y)-\phi(  h(X),Y):f\in
          B_R\right\}.
$$
 Using the obvious inequalities $\|f\|_\infty \leq R$,  $|y|\leq
M$ and $\| h\|_\infty\leq \|h\|_1$, from Assumption \ref{ASSUMPTION
1} it follows
   the inequalities
$$
            |\psi(Z)|\leq  H_\phi(R,M)+H_\phi(\|h\|_1,M)
$$
and
$$
           \mathbf
           E\psi^2
           \leq
            (H_\phi(R,M)+H_\phi(\|h\|_1,M))\mathbf E\psi.
$$
For $\psi_1,\psi_2\in\mathcal F_R$, it follows from Assumption
\ref{ASSUMPTION 2} that there exists a constant $C_4$ such that
\begin{eqnarray*}
          |\psi_1(Z)-\psi_2(Z)|
           &=&
          |\phi(f_1,Y)-\phi(f_2,Y)|\\
           &\leq&
           C_4|f_1(X)-f_2(X)|.
\end{eqnarray*}
We then get
$$
            \mathcal N(\mathcal
             F_R,\varepsilon)
             \leq
              \mathcal N(
             B_R,\varepsilon/C_4).
$$
According to Lemma \ref{COVERING ESTIMATE}, there holds
$$
           \log \mathcal N(\mathcal
             F_R,\varepsilon)
             \leq
              C_3k\log\left(\frac{4C_4R}{\varepsilon}\right).
$$
Employing  Lemma \ref{CONCENTRATION INEQUALITY} with
$B=c=H_\phi(R,M)+H_\phi(\|h\|_1,M)$ and
$$
       \mathbf E\psi=Q(f)-Q(h),\
       \frac1m\sum_{i=1}^m\psi(Z_i)=Q_m(f)-Q_m(h),
$$
we have, for every $\varepsilon>0$,
$$
              \sup_{f\in B_R}
             \frac{Q(f)-Q(h)-(Q_m(f)-Q_m(h))}{\sqrt{
             Q(f)-Q(h)+\varepsilon}}\leq\sqrt{\varepsilon}
$$
with confidence at least
$$
               1-
             \exp\left\{C_3k\log\left(\frac{4C_4R}{\varepsilon}\right)\right\}
             \exp\left\{-\frac{3m\varepsilon}{8\mathcal
             C(h,R,M)}\right\},
$$
where $\mathcal C(h,R,M)=(H_\phi(R,M)+H_\phi(\|h\|_1,M))$.

It follows from (\ref{boundedness of fk}) that $f_k\in B_R$ with
$R=C_5\log k$, then  with   confidence at least
$$
               1-
             \exp\left\{C_3k\log\left(\frac{C_6\log
             k}{\varepsilon}\right)\right\}
             \exp\left\{-\frac{3m\varepsilon}{8\mathcal
             C_1}\right\},
$$
there holds
\begin{eqnarray*}
              &&Q(f_k)-Q(h)-(Q_m(f_k)-Q_m(h))\\
              &&  \leq\sqrt{\varepsilon}(\sqrt{
             Q(f_k)-Q(h)+\varepsilon})\\
             &&\leq
             \frac12(Q(f_k)-Q(h))+\varepsilon,
\end{eqnarray*}
where  $\mathcal C_1= \mathcal C(h,C_5\log k,M)$.  Therefore, with
the same confidence, there holds
$$
         Q(f_k)-Q(h)\leq
         2(Q_m(f_k)-Q_m(h))+2\varepsilon.
$$
Since Assumptions \ref{ASSUMPTION 1} and \ref{ASSUMPTION 2} hold, it
follows from Theorem \ref{NUMERICAL CONVERGENCE} that for any
function $h\in\mbox{Span}(S)$, there holds
$$
       Q_m(f_k)-Q_m(h)\leq C(\|h\|_1^2+\log k)k^{-1},
$$
where $C$ is a constant depending only on $c_1,c_2$ and $C_1$.
Combining the last two inequalities yields that
$$
          \mathcal T\leq\varepsilon
$$
holds with
 at least
\begin{eqnarray*}
               &&1-
             \exp\left\{C_3k\log\left(\frac{C_6\log
             k}{\varepsilon}\right)\right\}\\
             &&\times
             \exp\left\{-\frac{3m\varepsilon}{8\mathcal
             C(h,C_5\log k,M)}\right\},
\end{eqnarray*}
where
$$
          \mathcal T=\frac{Q(f_k)-Q(h)- C(\|h\|^2_1+\log k)k^{-1}}{2}.
$$

For arbitrary $\mu>0$, there holds
\begin{eqnarray*}
             &&E_{\rho^m}(\mathcal T)
             =
             \int_{0}^\infty\mathbf P\{\mathcal T>\varepsilon\}d\varepsilon\\
             &\leq&
             \mu+\int_{\mu}^\infty
             \exp\left\{C_3k\log\frac{C_6\log k}{\varepsilon}-\frac{3m\varepsilon}{8\mathcal C_1}\right\}d\varepsilon\\
             &\leq&
             \mu+\exp\left\{-\frac{3m\mu}{8\mathcal C_1}\right\}
             \int_\mu^\infty\left(\frac{C_6\log k}{\varepsilon}\right)^{C_3k}d\varepsilon\\
             &\leq&
             \mu+\exp\left\{-\frac{3m\mu}{8\mathcal C_1}\right\}\left(\frac{C_6\log
             k}{\mu}\right)^{C_3k}\mu.
\end{eqnarray*}
By setting $\mu=\frac{\mathcal C_1C_3k(\log m+\log k)}{3m}$, direct
computation yields
$$
                   \mathbf E(\mathcal T)\leq \frac{2\mathcal C_1C_3k(\log m+\log
                    k)}{3m}.
$$
That is,
\begin{eqnarray*}
         &&\mathbf E\{  Q(f_k)-Q(h)\}\\
         &\leq&
          C(\|h\|_1^2+\log k)k^{-1}+ \frac{4\mathcal C_1C_3k(\log m+\log
                    k)}{3m},
\end{eqnarray*}
which finishes the proof Theorem \ref{LEARNING RATE BOOSTING}.
\end{IEEEproof}

\section{Conclusion and further discussions}

In this paper, we proposed a new idea to conquer the slow numerical
convergence rate problem of boosting and then develop a new variant
of boosting, named as the re-scale boosting (RBoosting). Different
from other variants such as the $\varepsilon$-boosting, RTboosting,
RSboosting that control the step-size in the linear search step,
RBoosting focuses on alternating the direction of linear search via
implementing a re-scale operator on the composite estimator obtained
by the previous iteration step. Both theoretical and experimental
studies illustrated that RBoosting outperformed the original
boosting and performed at least comparable to other variants of
boosting. Theoretically, we proved that the numerical convergence
rate of RBoosting was almost optimal in the sense that it cannot be
essentially improved. Using this property, we then deduced a fairly
tight generalization error bound of RBoosting, which was a new
``record'' for boosting-type algorithms. Experimentally, we showed
that for a number of numerical experiments, RBoosting outperformed
boosting, and
  performed at least the second best of all   variants of
boosting. All these results implied that RBoosting was an reasonable
improvement of Boosting and the idea ``re-scale'' provided a new
direction to improve the performance of boosting.

To stimulate more opinions from others on RBoosting, we present the
following two remarks at the end of this paper.

\begin{remark}
Throughout the paper, up to the theoretical optimality, we can not
provide any essential advantages of RBoosting in applications, which
makes it difficult to persuade the readers to use RBoosting rather
than other variants of boosting. We highlight that there may be two
merits of RBoosting in applications. The first one is that, due to
the good theoretical behavior, if the parameters of RBoosting are
appropriately selected, then RBoosting may outperform other
variants. This conclusion was partly verified in our experimental
studies in the sense that for all the numerical examples, RBoosting
performed at least the second best. The other merit is that,
compared with other variants, RBoosting cheers a totally different
direction to improve the performance of boosting. Therefore,    it
paves a new way to understand and improve boosting. Furthermore, we
guess that if the idea of the ``re-scale'' in RBoosting and
``regularization'' in other variants of boosting are synthesized to
develop a new boosting-type algorithm, such as the re-scale
$\varepsilon$-boosting, re-scale RTboosting, then the performance
may be further improved. We will  keep working on this    issue and
report our progress in a future publication.
\end{remark}

\begin{remark}
According to the ``no free lunch'' philosophy, all the variants
improve the learning performance of boosting at the cost of
introducing   additional parameters, such as the truncated parameter
in RTboosting, regularization parameter in RSboosting, $\varepsilon$
in $\varepsilon$-Boosting, and shrinkage degree in RBoosting. To
facilitate the use of these variants, one should also present
strategies to   select such parameters. In particular, Elith et al.
\cite{Elith2008} showed that $ 0.1$ is a feasible choice of
$\varepsilon$ in $\varepsilon$-Boosting; B\"{u}hlmann and Hothorn
\cite{Buhlmann2007} recommended the selection of $0.1$ for the
regularization parameter in RSboosting; Zhang and Yu
\cite{Zhang2005} proved that $\mathcal O(k^{-2/3})$ is a good value
of the truncated parameter in RTboosting. One may naturally ask:
how to select the shrinkage degree $\alpha_k$ in RBoosting?
 This is a good
question and we find a bit headache to answer it. Admittedly, in
this paper, we do not give any essential suggestion to practically
attack this question.
 In fact,
 $\alpha_k$ plays an important role in RBoosting. If
$\alpha_k$ is too small, then RBoosting performs similar as the
original boosting, which can be regarded as a special RBoosting with
$\alpha_k=0$. If $\alpha_k$ is too large, an extreme case is
$\alpha_k$ close  to $1$, then the numerical convergence rate of
RBoosting is also slow. Although  we theoretically present some
values of the $\alpha_k$, the best one in applications, we think,
should be selected via some model selection strategies. We leave
this important issue into a future study \cite{Xu2015}, where the
concrete role of the shrinkage degree will be revealed.
\end{remark}

\section*{Acknowledgement}
The research was supported by the National Natural Science
Foundation of China (Grant No. 11401462) and the National 973
Programming (2013CB329404).

\end{document}